\documentclass[sn-nature]{sn-jnl}

\usepackage[table]{xcolor}
\usepackage{graphicx}%
\usepackage{bm}
\usepackage{multirow}%
\usepackage{amsmath,amssymb,amsfonts}%
\usepackage{amsthm}%
\usepackage{mathrsfs}%
\usepackage{xcolor}%
\usepackage{textcomp}%
\usepackage{manyfoot}%
\usepackage{booktabs}%
\usepackage{algorithm}%
\usepackage{algorithmicx}%
\usepackage{algpseudocode}%
\usepackage{listings}%
\usepackage{appendix}

\theoremstyle{thmstyleone}%
\theoremstyle{thmstyletwo}%

\theoremstyle{thmstylethree}%

\begin{document}

\title[Article Title]{Automated discovery of symbolic laws governing skill acquisition from naturally occurring data}


\author[1,2,3]{\fnm{Sannyuya} \sur{Liu}}

\author[1,2]{\fnm{Qing} \sur{Li}}

\author*[1,2]{\fnm{Xiaoxuan} \sur{Shen}}\email{shenxiaoxuan@ccnu.edu.cn}

\author*[1,2]{\fnm{Jianwen} \sur{Sun}}\email{sunjw@ccnu.edu.cn}

\author*[1,2,3]{\fnm{Zongkai} \sur{Yang}}\email{zkyang027@ccnu.edu.cn}

\affil[1]{\orgdiv{National Engineering Research Center of Educational Big Data}, \orgname{Central China Normal University}, \orgaddress{\city{Wuhan}, \postcode{430079}, \country{China}}}
\affil[2]{\orgdiv{Faculty of Artificial Intelligence in Education}, \orgname{Central China Normal University}, \orgaddress{\city{Wuhan}, \postcode{430079}, \country{China}}}
\affil[3]{\orgdiv{National Engineering Research Center for E-learning}, \orgname{Central China Normal University}, \orgaddress{\city{Wuhan}, \postcode{430079}, \country{China}}}


\abstract{Skill acquisition is a key area of research in cognitive psychology as it encompasses multiple psychological processes. The laws discovered under experimental paradigms are controversial and lack generalizability. This paper aims to unearth the laws of skill learning from large-scale training log data. A two-stage algorithm was developed to tackle the issues of unobservable cognitive states and algorithmic explosion in searching. Initially a deep learning model is employed to determine the learner's cognitive state and assess the feature importance. Subsequently, symbolic regression algorithms are utilized to parse the neural network model into algebraic equations. Experimental results show the algorithm can accurately restore preset laws within a noise range in continuous feedback settings. When applied to Lumosity training data, the method outperforms traditional and recent models in fitness terms. The study reveals two new forms of skill acquisition laws and reaffirms some previous findings.}

\maketitle

\section{Introduction}

Skill acquisition theory explains how individuals develop skills through practice and experience, emphasizing the processes of acquiring, refining, and automating skills \cite{SA1}. A key concept in this theory is deliberate practice, which involves focused efforts to improve specific areas of performance \cite{SA2}. This theory has had a significant impact on fields like sports, music, education, and programming, offering valuable insights into effective skill acquisition and improvement across different domains \cite{tabibian2019enhancing}.

Skill acquisition is a complex psychological process involving practice \cite{practice1,practice2,practice3}, memory \cite{forgetting1,forgetting2,forgetting3}, and skill transfer \cite{transfer1,transfer2}. Researchers use symbolic representation methods, like algebraic equations, to describe experimental phenomena and identify patterns. Establish cognitive theoretical models based on observed experimental patterns. Instance theory \cite{instance1,instance2} explains the power law \cite{anderson1982acquisition} of practice by suggesting that skill improvement results from accumulating and refining relevant instances during practice repetitions. Similarly, there are also theories such as multi-phase theory \cite{stage1,stage2} and parallel distributed processing models \cite{pdp1,pdp2}, along with cognitive architectures like SOAR \cite{Soar} and ACT-R \cite{ACT-R1,ACT-R2}. Symbolic patterns are essential for developing cognitive theories, but their complexity and controversy highlight the need for further study in understanding human behavior.

Traditional psychological research often derives symbolic rules from controlled experiments, leading to varying conclusions due to different perspectives. For instance, the law of practice, initially believed to follow the power law by many researchers \cite{practice3,anderson1982acquisition}, but some studies suggesting there are better symbol representations, like exponential or APEX functions \cite{practice2}, in terms of data fitting \cite{practice1}. Besides, biases and limitations in experimental designs can hinder result replication and generalization \cite{NODS1}. Jenkins’ tetrahedral model \cite{tetrahedral} emphasizes considering subjects, encoding activities, events, and retrieval conditions in memory experiments, indicating that findings from controlled experiments may apply only to specific conditions. Further research is required to evaluate the robustness and generalizability of findings across diverse contexts.

Psychologists have turned to analyzing human behavior patterns using naturally occurring data sets (NODS) to overcome challenges in traditional psychological research. This approach, known as the "big data" paradigm, complements lab experiments by extracting psychological patterns from large real-world datasets \cite{NODS1}. While lacking experimental control, NODS allow for statistical separation of factors and uncover novel patterns that are not commonly investigated in previous research. The use of empirical data from realistic settings enhances the applicability of conclusions, and the vast scale of data leads to more robust findings \cite{NODS1}. The emergence of online training systems has provided learners with opportunities to improve skills through practice sessions, generating extensive training log data with valuable behavioral patterns related to skill acquisition. It presents opportunities to develop theories and models using the "big data" paradigm.

Symbolic regression, a machine learning technique, aims to find mathematical expressions that best fit a dataset by combining operations, functions, and variables \cite{udrescu2020ai}. It is valuable for uncovering data patterns and inspiring further research \cite{wang2023scientific}. As a search-based algorithm, symbolic regression typically provides optimal results within the search space, making it suitable for identifying skill acquisition patterns and applications in behavioral sciences. While it has been widely used in physics \cite{SRA1,SRA2,SRA3,SRA5}, chemistry \cite{SRA4,SRA6}, and materials science \cite{SRA7,SRA8}. However, the complexity of training log data, including scale, dimensionality, and latent states, poses challenges for applying symbolic regression algorithms directly. Traditional algorithms like genetic algorithms struggle to find optimal symbolic models due to the exponential growth in solution space with variables and operators \cite{udrescu2020ai}. The numerous uncontrolled variables and factors affecting skill acquisition in NODS hinder direct optimization of symbolic models. Moreover, the skill mastery of learner is unobservable, present obstacles for current data analysis algorithms to uncover symbolic rules related to skill acquisition from training log data.

Deep learning models, or deep neural networks, have become popular in machine learning for their ability to effectively analyze data in high-dimensional space using the back-propagation algorithm \cite{lecun2015deep}. These models excel at capturing complex data patterns, enabling accurate estimation of learner skill mastery. However, a drawback is their black-box nature, making it hard for humans to interpret the decision-making process behind the fitted data patterns.

\begin{figure}[th!]%
	\centering
	\includegraphics[width=\textwidth]{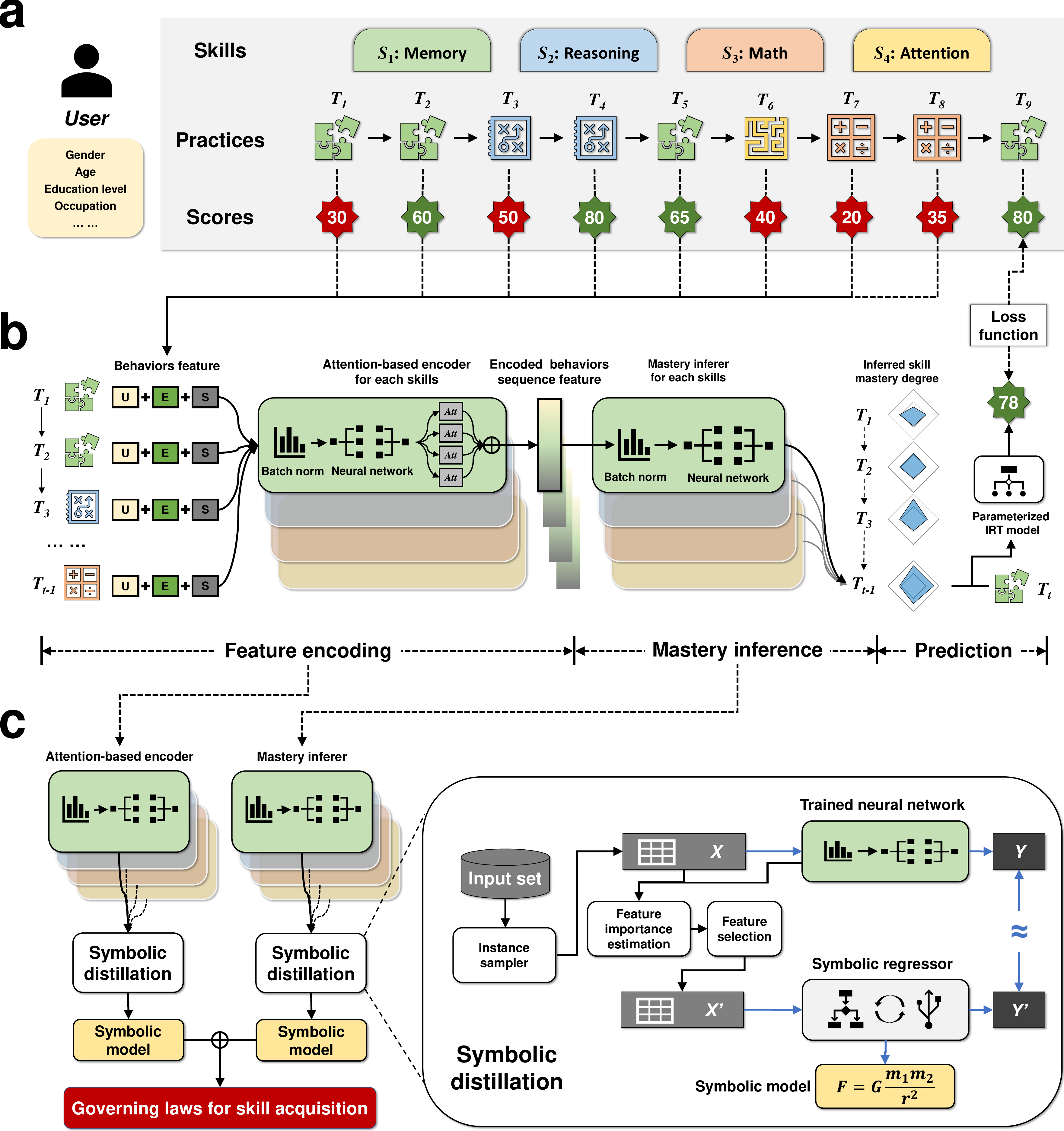}
	\caption{\textbf{Overall model architecture diagram.} A two-stage model is proposed for the automated discovery of symbolic laws that govern skill acquisition. Specifically, (a) provides a toy example of training log data. The example illustrates the relevant information of nine practice sessions for a specific learner, where each record containing five main elements: learner, practice, skill, score, and time. (b) depicts the constructed deep regressor. It consists of three main modules: feature encoding, mastery inference, and score prediction. The model takes the output of the previous n practice sessions and predicts the score of the next practice session. The proposed model follows the autoregressive paradigm to establish the optimization objective of the model. (c) describes the process of extracting symbolic governing laws from the trained deep regressor. First, we propose the symbolic distillation method, which represents the black-box neural network model as the closest symbolic representation. Secondly, for a trained deep regressor, it has already embedded hidden patterns from the data into the model. The symbolic distillation method is utilized to interpret the various modules of the deep regressor and fuse them together to obtain symbolic governing laws.}\label{fig1}
\end{figure}

This paper proposes a method to automatically discover hidden patterns in skill acquisition from large training log data. A two-stage model is presented, combining deep learning and symbolic regression algorithms to reveal the fundamental principles of skill acquisition. This approach guarantees accurate fitting and interpretability by extracting key patterns from complex training logs and presenting them in algebraic equation, enhancing understanding. Figure 1 illustrates the overall framework of the proposed method.

The first stage of the proposed two-stage model focuses on developing a deep learning regressor to accurately fit large-scale online training log data, shown in Fig. 1b. This regressor aims to estimate crucial latent variables, such as behavior feature encoding and user skill proficiency, and to fit key mapping functions precisely. Drawing inspiration from the successful transformer model \cite{transformer} for sequence modeling, we devised a transformer-like model. Initially, it evaluates the importance of behaviors in each skill training by utilizing static behavior features and generates encoded sequential behavior features. Subsequently, a non-linear neural network model computes the user’s skill mastery state based on the encoded features. Finally, the skill proficiency level is linked to the training score using the three-parameter item response theory (IRT) model \cite{IRT1,IRT2,IRT3,IRT4}, a well-established cognitive diagnostic model. The model is trained using an autoregressive approach that predicts future training outcomes based on historical training sequences, enabling a comprehensive exploration of users’ behavior patterns during the training process.

In the second phase of the study, symbolic regression was used to extract core patterns from a trained deep learning regressor into algebraic equations, resulting in symbolic rules (Fig. 1c). Symbolic distillation was developed to extract these patterns from the neural network by creating a smaller pattern extraction dataset from a large training dataset. By combining this with a gradient-based neural network feature importance estimation algorithm, crucial features were selected, reducing the solution space for symbolic regression. The symbolic model was optimized by minimizing prediction error between the symbolic model and the trained neural network using the pattern extraction dataset. Symbolic distillation techniques were then applied to behavior importance estimation and mastery estimation networks post deep learning model fitting. Finally, symbolized rules were consolidated to derive skill acquisition rules.

Thus far, we have outlined the rationale and general process of the proposed method. A detailed description of the model will be presented in sections 4.1 and 4.2 of the Method section. The effectiveness and feasibility of the method are confirmed through experiments on simulated data in the Results section. Following this, the method is applied to the Lumosity dataset for large-scale skill training to reveal hidden patterns and regularities in the data.

\section{Results}\label{sec2}

\subsection{Model validation on simulated data}
Validating patterns from data in skill acquisition is challenging, due to their contentious nature. So the model validation is addressed through simulated experiments. The validation aims to demonstrate the algorithm's ability to restore data patterns, enhancing reliability in real-world applications. Algebraic equations from skill acquisition theories are used as preset governing laws, shown in Table 1. Then, simulated learners, exercises and training logs are generated with skill mastery and practice scores computed by preset governing laws. The study tests if the algorithm can restore these laws using the simulated data only. Extended Data Figure 2 provides details on simulated data generation, evaluation process, and sample representation. Further details on the process of generating simulated data, including data structure and parameter settings, will be elaborated in the Method section.

\begin{table}[h]
	\caption{The predetermined practice rules - the relationship function between practice times and skill mastery. Here, $N_1$ and $N_2$ refer to the number of times the learner practiced skill 1 and skill 2, respectively. $\alpha$ and $\beta$ are the exercise rate parameters, $\mu$ is the mutualism factor, $\gamma$ represents the forgetting rate. {In the simulation, we only establish mutualism for skill 1, meaning that practicing skill 2 aided in mastering skill 1. Therefore, the number of practice sessions for the skill 1 considering mutualism is defined as $((1-\mu) \cdot N_1+ \mu \cdot N_2)$.}}
	\begin{tabular}{@{}llc@{}}
		\toprule
		SETTINGS & SKILLS  & GOVERNING LAWS\\
		\midrule
		SETTING 1 & SKILL 1: Linear   & $\beta \cdot \alpha \cdot ((1-\mu) \cdot N_1+ \mu \cdot N_2)- \gamma \cdot (N_1+N_2)$  \\
		Linear+ Exponential    & SKILL 2: Exponential   & $\beta \cdot \exp(\alpha \cdot N_2)-\gamma\cdot(N_1+N_2)$  \\
		\midrule
		SETTING 2    & SKILL 1: Exponential   & $\beta \cdot \exp(\alpha \cdot ((1-\mu) \cdot N_1+ \mu \cdot N_2))-\gamma\cdot(N_1+N_2)$  \\
		Exponential + Power    & SKILL 2: Power   & $\beta \cdot {N_2}^\alpha-\gamma\cdot(N_1+N_2)$  \\
		\midrule
		SETTING 3    & SKILL 1: Power   & $\beta \cdot {((1-\mu) \cdot N_1+ \mu \cdot N_2)}^\alpha-\gamma\cdot(N_1+N_2)$   \\
		Power + Linear    & SKILL 2: Linear   & $\beta \cdot \alpha \cdot N_2- \gamma \cdot (N_1+N_2)$   \\
		\botrule
	\end{tabular}
	\label{table1}
\end{table}

\begin{figure}[th!]%
	\centering
	\includegraphics[width=1\textwidth]{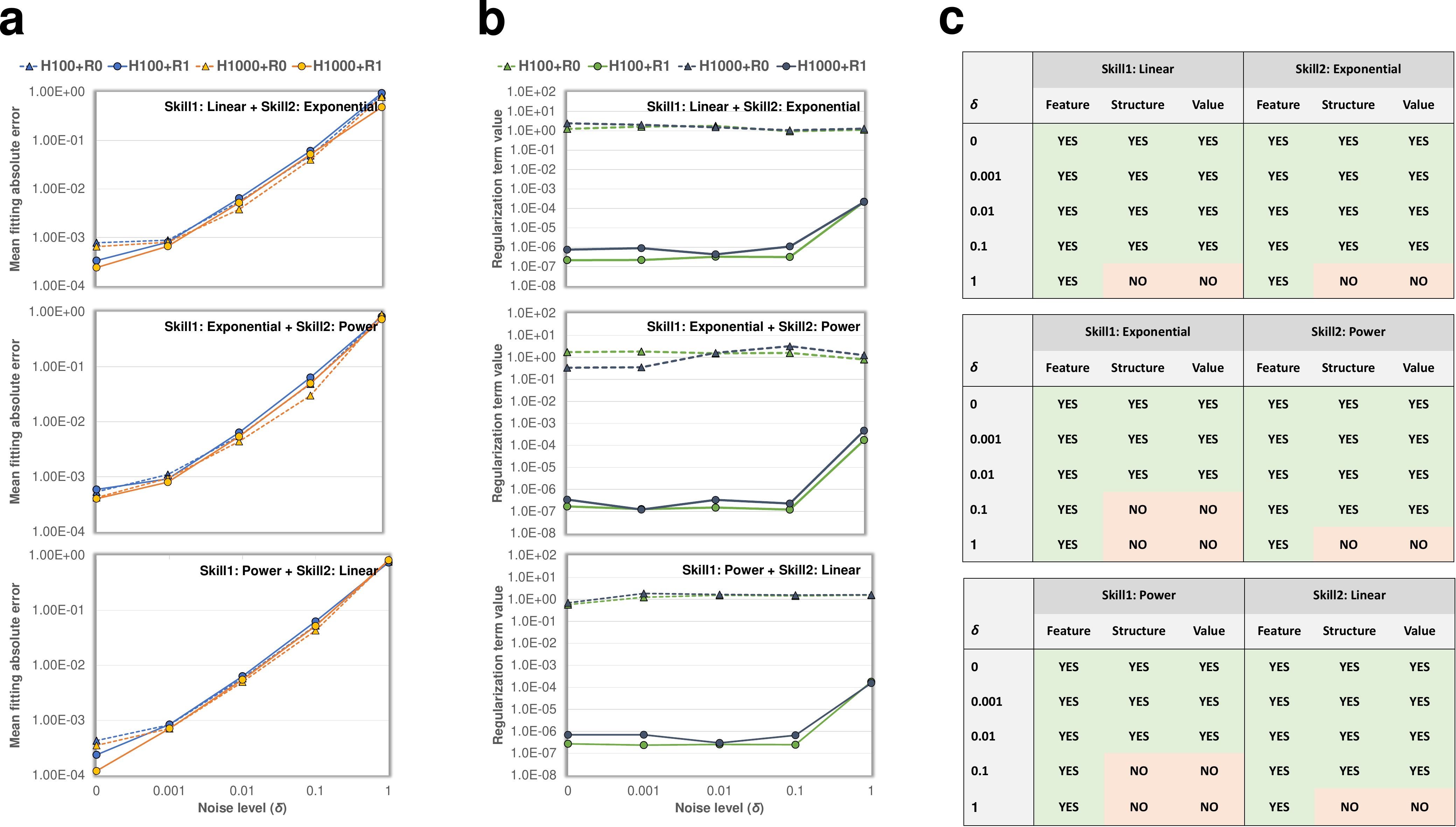}
	\caption{\textbf{Results of the simulated data experiment.} The results of the model experiments on simulated data. Specifically, it presents more detailed analyses of (a) the fitting degree of the deep learning fitter under different parameter settings, (b) the numerical analysis of regularization terms under different parameter settings, and (c) the analysis of the degree of accuracy in reproducing governing laws. {Feature, Structure, and Value are introduced as three dimensions to evaluate the degree of formula restoration. Feature measures whether the variables in the restored formula match those in the assumed formula. Structure indicates whether the form and structure of the restored formula align with the assumed formula. Value represents whether the parameters in the formula match those in the assumed formula. In general, if the Feature, Structure, and Value are correctly restored, the formula can be considered completely restored and accurate.} "Yes" indicates that the algorithm has successfully restored the predetermined pattern, whereas "No" indicates its failure to do so.}\label{fig2}
\end{figure}

First, we investigated the fitting of the deep regressor to the simulated data. We created four deep learning regressors with different parameters: H100+R0, H100+R1, H1000+R0, and H1000+R1. Here, H represents the number of hidden neurons per layer in the model \cite{capabilities1,capabilities2}, with H set to 100 and 1000 to simulate small and large fitting abilities. R indicates the use of regularization for attention value, with R=1 or 0 representing with or without regularization in the deep regressor's optimization objective. Regularization can simplify the restoration equation, improve model readability, but may reduce fitting ability. We trained the four regressors for 5000 epochs on simulated datasets at five error levels ({$\delta\in \{0,0.001,0.01,0.1,1\}$}). The average fitting absolute error and regularization term's value are plotted in Fig. 2a and Fig. 2b, respectively.

Fig. 2a shows that models with more hidden neurons (H1000+R0 and H1000+R1) perform slightly better than those with fewer neurons (H100+R0 and H100+R1), consistent with prior neural network studies \cite{NN1}. The model with regularization (R1) has slightly lower fitting accuracy than the model without regularization (R0), but the difference is not significant. In Fig. 2b, the regularization term for the R1 model is much smaller (3 to 7 orders of magnitude) than that of the R0 model. Meanwhile, the H1000+R1 model has a slightly lower regularization term than the H100+R1 model. Overall, the R1 model shows minimal deviation in fitting error compared to the non-regularized R0 model, while vastly reducing the regularization term value. The H1000 model outperforms the H100 model in fitting performance and regularization term value. Therefore, the H1000+R1 model is chosen as the deep learning regressor for symbolic distillation and analysis of governing laws in subsequent experiments.

Secondly, we analyze the extent to which the model reproduces the predetermined governing laws. Based on the trained H1000+R1 model, the process of symbolization distillation and extraction of governing laws is performed, and the resulting governing laws are compared with the predetermined model to assess the degree of model restoration. 

In deep learning regressor, we used five basic features from two aspects: exercise characteristics (E) and scheduling information (S). The E feature involves one-hot encoding of exercise correlations with SKILL 1, SKILL 2, and a general factor \cite{Gf1,Gf2}, treated as a special skill \cite{dataset1}. The S feature includes Count (number of skill practices) and Interval (time between practice sessions). In the H1000+R1 model, the E feature weight is set to 1 due to its extremely low regularization term value (1e-7 to 1e-4), resulting in encoded features Sum\_G, Sum\_S1, and Sum\_S2, representing historical practice times and frequency for corresponding skills. 

Model evaluation considers Feature, Structure, and Value dimensions. Feature dimension focuses on identifying the importance of Sum G, Sum S1, and Sum S2. Structure dimension assesses the correctness of the algebraic equation from symbolic regression, allowing some error in parameter values. Value dimension checks if the restored equation falls within an acceptable range of the true value, with an error threshold of less than 1\%. These dimensions are interconnected in the evaluation process.

Simulation experiments were conducted to test the restoration of governing laws on simulated datasets across three settings and four noise levels. Results are shown in Fig. 2c, with some feature importance results in Supplementary Figure 1. The proposed method effectively restored governing laws for preset skills within a noise range ($\delta < 0.1$). At $\delta = 0.1$, the model reproduced equations for SKILL 2 and Linear setting equation for SKILL 1 but faced challenges with Power and Exponential settings due to errors causing ambiguity in function determination. This challenge lessened with smaller errors. At $\delta = 1$, the model identified the core features but struggled with equation structure due to many noise values exceeding the rating range [0, 1]. Results under alternative error settings in Supplementary Figure 2 showed similar results. The results indicate that the proposed method is capable of accurately reconstructing the predefined equations within a certain range of error in three different settings, demonstrating its potential applicability in real-world scenarios.

\subsection{Model application on large-scale real-world cognitive training data }

In this section, we attempted to apply the method proposed in this paper to large-scale real-world datasets to discover patterns from real data and provide evidence for skill acquisition theory. For the application, we selected Lumosity as the validation platform \cite{lumos}. A more detailed description of the dataset, data filtering, preprocessing methods, etc., will be provided in the Method section.

Model fitting analysis was conducted to assess the performance of a deep learning regressor on actual data and perform model selection. Six models were evaluated: H100+R0, H100+R1, H500+R0, H500+R1, H1000+R0, and H1000+R1. The naming convention follows that of the simulation experiment, with H indicating the number of hidden neurons per layer and R indicating the inclusion of a regularization term in the optimization objective. Each model underwent 50 training epochs, with the mean fitting absolute error and regularization term values plotted over epochs (Extended Data Figure 1a and 1b). Results showed convergence of all models with error values ranging from 0.086 to 0.088. The R1 model optimized the regularization term to a value near 1e-5, approximating 1 for attention value. The H1000+R1 model, exhibiting the best fitting performance, was selected for further experiments.

\begin{figure}[th!]%
	\centering
	\includegraphics[width=0.9\textwidth]{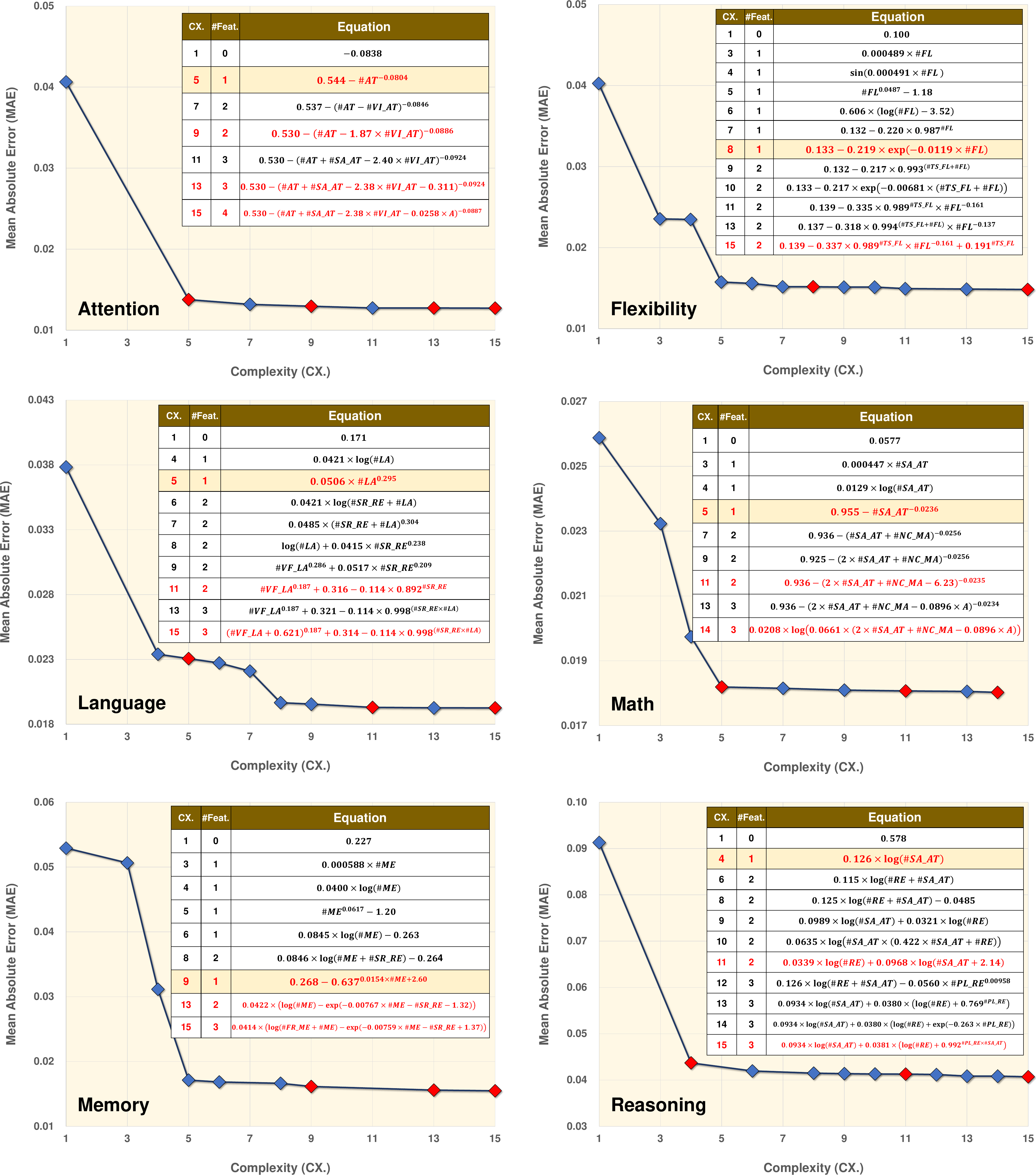}
	\caption{\textbf{The skill acquisition patterns discovered by the proposed method.} The symbolic governing laws discovered by the model on the Lumosity dataset. The symbol regression results for six related skills and their corresponding complexities (CX.$<$15) are presented, including algebraic equations and their mean absolute error (MAE). The best-fitting equation with the same number of features is marked in red.}\label{fig3}
\end{figure}

To further assess the deep learning regressor's performance, we randomly sampled data and examined the model's absolute fitting error across different skills, shown in Extended Data Figure 1c. The error distributions for Attention, Flexibility, Reasoning, and Language skills were similar. However, Memory had the lowest average error, while Math had the highest, approximately 50\% more than other skills. This indicates that Math skill training may involve more complexity or unmodeled key variables. Additionally, the proportion of training log data for each skill relative to the total data is depicted in Extended Data Figure 1d, showing relatively low proportions for Math, Reasoning, and Language.

Feature importance analysis was conducted for the H1000+R1 deep learning regressor to identify key factors influencing skill mastery. Age and education level were found to be important for all skills, indicating their impact on skill acquisition. Game play can improve reaction speed and attention, while education level reflects initial abilities, particularly for Math and Language. Practice times, represented by "\#" symbols, were crucial for skill acquisition. Additionally, certain subskills, such as SR\_RE (All the features used are explained in Extended Data Figure 2) for Reasoning, were found to influence Language mastery, suggesting mutualism. These findings align with previous research \cite{mvr1,mvr2}.

Analysis of symbolic governing laws is shown in Figure 3, presenting optimal equations for six skills at various levels of complexity. Symbol regression algorithms perform better with more complex equations, but fitting error for skills does not decrease significantly beyond a certain complexity threshold. We used the number of features in equations as a metric and focused on identifying the best fitting equations for each skill with different feature quantities (highlighted in red in Fig. 3). Skills showed good fitting performance with one feature, indicating a core pattern. Increasing complexity did not improve fitting significantly. Therefore, we focused on the best fitting equation with one feature (highlighted in yellow in the tables of Fig. 3).

\subsection{Exploring auto-discovered patterns in real-world datasets}
To analyze the identified patterns, we compared them with classical Power and Exponential laws using data from Lumosity on coding features and skill mastery across six skills. Parameters in these equations were optimized, and goodness of fit was evaluated using the $R^2$ metric \cite{r2}. Results are summarized in Table 2, with visual representations for pattern assessment. A sample of five learners had their skill mastery estimation curves compared with outcomes from the deep regressor, shown in Extended Data Figure 4.

\begin{table}[h]
	\caption{The comparisons on the goodness of fit between the discovered laws and the classical Power law and Exponential law for six skills using the Lumosity dataset. $R^2$ is introduced as a measure of goodness of fit, which represents the proportion of the target variable's variance explained by the model and ranges from 0 to 1. A higher value indicates a better fit. The model with the highest goodness of fit within the same skill is bolded, while the model ranked second is underlined.}
	\begin{tabular}{c|ccc}
		\toprule
		Skill& Law Type &{Equation} & $R^2\uparrow$ \\ \midrule
		\multirow{3}{*}{\textbf{Attention}} & Power & \tiny $0.340-0.821\times\#AT^{-0.114}$ & \underline{0.850}  \\
		& Exponential & \tiny $-0.049-0.194\times exp(-0.005\times\#AT)$& 0.809 \\
		& Discovered & \tiny {$0.544-{\#AT}^{-0.081}$} & \textbf{0.856} \\\midrule
		\multirow{3}{*}{\textbf{Flexibility}}& Power& \tiny$-0.376+0.231\times\#FL^{0.139}
		$& 0.841   \\
		& Exponential& \tiny$0.133-0.219\times exp(-0.011\times\#FL)$& \textbf{0.850} \\
		& Discovered&\tiny$0.133-0.219\times exp(-0.011\times\#FL)
		$& \textbf{0.850} \\\midrule
		\multirow{3}{*}{\textbf{Language}}& Power & \tiny$ 0.026+0.033\times\#LA^{0.354 }$ & \textbf{0.527}\\
		&Exponential& \tiny$ -0.103+0.214\times exp(0.003\times\#LA) $ & 0.477 \\
		& Discovered &\tiny $0.051\times\#LA^{0.295}$ &\underline{ 0.525 }\\\midrule
		\multirow{3}{*}{\textbf{Math}}& Power & \tiny$ -0.139+0.113\times\#MA^{0.121 }$& \underline{0.202} \\
		&Exponential & \tiny$ -0.054+0.086\times exp(0.002\times\#MA) $& 0.115 \\
		& Discovered & \tiny$ 0.955-\#SA\_AT^{-0.024 }$ & \textbf{0.278}  \\\midrule
		\multirow{3}{*}{\textbf{Memory}}& Power & \tiny$ -0.020+0.020\times\#ME^{0.427} $& 0.829 \\
		&Exponential & \tiny$ 0.269-0.279\times exp(-0.006\times\#ME) $& \underline{0.885} \\
		& Discovered & \tiny$ 0.268-0.637^{0.015\times\#ME+2.60 }$& \textbf{0.891}\\\midrule
		\multirow{3}{*}{\textbf{Reasoning}}& Power &\tiny$ -0.133+0.376\times\#RE^{0.159} $ & \underline{0.605 }\\
		&Exponential& \tiny$ -0.041+0.513\times exp(-0.003\times\#RE) $ & 0.265 \\
		& Discovered&\tiny$  0.126\times log(\#SA\_AT) $ & \textbf{0.698} \\
		\botrule
	\end{tabular}\label{table2}
\end{table}

The results in Table 2 show some interesting phenomenon. Firstly, the Power law and Exponential law had varying fitting performances on different skills. The Power law fit well for Attention, Language, Math, and Reasoning skills, while the Exponential law was better for Flexibility and Memory skills. This discrepancy is consistent with previous research findings \cite{practice1,practice2}. Secondly, our method generally outperformed the traditional laws, especially for Attention, Flexibility, Language, and Reasoning skills. Symbolic regression equations provided better fitting results in most cases. Minor parameter variations caused slight differences, highlighting the effectiveness of our approach without prior knowledge of traditional skill acquisition laws. Lastly, our method revealed new patterns for Math and Memory skills, deviating from conventional laws and showing improved fitting performance. This demonstrates the potential of our method to uncover novel patterns and offer valuable insights for researchers, emphasizing a data-driven approach.

Upon examining the equations discovered, various practice patterns emerged. Logarithmic law was identified in some equations, a type not previously mentioned in research. Additionally, two inverse forms of power law were observed for Attention ($0.544-{\#AT}^{-0.0804}$) and Language ($0.0506\times{\#LA}^{0.295}$), with parameters that exhibit opposite effects. The absence of general factor theory in this study suggests it may not significantly impact skill acquisition. Skill forgetting was not clearly observed too. The main factors influencing Language and Math were found to be the frequency of attention selection practice rather than other practice frequency. This highlights the importance of attention in game-based skill improvement. However, this phenomenon may imply improving Language and Math skills through gamified practice may pose challenges. The experimental results suggest that our algorithm has the capability to discover the pattern forms previously observed by humans without any prior knowledge, and at the same time, to identify unknown pattern forms with better fitting. This fully demonstrates the application potential and value of the proposed method, which can effectively serve researchers in providing insights as data analysis tools.

\begin{table}[h]
	\centering
	\caption{The results of the goodness of fit ($R^2$) and the Bayesian information criterion (BIC) of all comparative models on the Lumosity dataset are reported. All reported results are the average of 20 independent experiments. The overall results of the dataset and the results for each skill are separately reported. A t-test for significance testing was conducted between the results of the best model and the second-best model, and the corresponding \textit{p}-values are reported. During the implementation of the t-test, two-sided and heteroscedasticity  settings are deployed, and  the bootstrap technique is employed to enhance the reliability of our findings. In each independent experiment, we randomly sampled 50\% of the dataset and subsequently repeated this iterative procedure five times. Results with significant differences (\textit{p}-value$<$0.01) are indicated with an asterisk (*). The results of the best-performing model are highlighted in bold, while the second-best model is marked with an underline. $\uparrow$ represents the larger the better, while $\downarrow$ represents the smaller the better. All BIC values are expressed in units of $10^7$.}
	\centering
	\begin{tabular}{cccccccc}
		\toprule
		$ R^2 \uparrow$ & Overall & Attention & Flexibility & Language & Math & Memory & Reasoning \\ \midrule
		AFM & 0.255  & 0.254  & 0.362  & 0.305  & 0.170  & 0.162  & 0.298  \\ 
		LGM & 0.370  & 0.380  & 0.523  & 0.519  & 0.297  & 0.225  & 0.426  \\ 
		BLM & 0.439  & 0.459  & 0.606  & 0.573  & 0.448  & 0.267  & 0.462  \\ 
		TLM & 0.486  & 0.503  & 0.624  & 0.613  & 0.477  & 0.301  & 0.593  \\ 
		TLMF & 0.471  & 0.475  & 0.614  & 0.624  & 0.483  & 0.287  & 0.592  \\ 
		IM & \underline{0.503}  & \underline{0.516}  & \underline{0.633}  & \underline{0.629}  & \underline{0.517}  & \underline{0.313}  & \underline{0.599}  \\ \midrule
		ADM & \textbf{0.627*} & \textbf{0.646*} & \textbf{0.645*}  & \textbf{0.641*}  & 0.423  & \textbf{0.643*}  & \textbf{0.628*}  \\ 
		\textit{p}-value & 8.93E-52 & 2.12E-45 & 2.76E-05 & 5.59E-08 & 1.86E-39 & 2.14E-50 & 8.43E-20 \\ \botrule
		\toprule
		$BIC \downarrow$ & Overall & Attention & Flexibility & Language & Math & Memory & Reasoning \\ \midrule
		AFM & -3.814 & -3.951 & -3.983 & -4.017 & -3.486 & -3.601 & -4.112 \\ 
		LGM & -4.020 & -4.179 & -4.340 & -4.469 & -3.691 & -3.697 & -4.361 \\ 
		BLM & -4.217 & -4.400 & -4.624 & -4.691 & -4.058 & -3.812 & -4.505 \\ 
		TLM & \underline{-4.264} & \underline{-4.440} & -4.698 & \underline{-4.821} & -4.059 & \underline{-3.815} & \underline{-4.760} \\ 
		TLMF & -4.174 & -4.321 & -4.541 & -4.684 & -3.986 & -3.736 & -4.719 \\ 
		IM & -4.242 & -4.413 & \underline{-4.706} & -4.725 & \underline{-4.080} & -3.771 & -4.741 \\ \midrule
		ADM & \textbf{-4.779*} & \textbf{-4.979*} & \textbf{-4.820*} & \textbf{-4.912*} & \textbf{-4.135*} & \textbf{-4.771*} & \textbf{-4.882*} \\ 
		\textit{p}-value & 5.59E-63 & 2.83E-46 & 7.59E-24 & 1.13E-15 & 7.97E-15 & 2.44E-69 & 3.08E-25 \\ \botrule
	\end{tabular}
\end{table}\label{table3}

Our study model focuses on analyzing learners' performance throughout the learning process, a key research area. To evaluate our model's performance, we compare it with various recent learning models using large-scale data. The comparative models include baseline learning model (BLM), two-timescale learning model (TLM), two-timescale learning model with forgetting (TLMF), and interactive model (IM) from \cite{learningC1}. In addition, we consider the classical additive factor model (AFM) \cite{learningC2} and the latest learning growth model (LGM) \cite{learningC3} for learner growth modeling. Our model, the Auto-discovered model (ADM), combines discovered laws with the 3-parameter IRT model for predicting learning performance. Unlike the comparison models that have unique parameters for each learner, our model is a general learning model where all learners share the same parameter set for learning patterns. We use two evaluation metrics: $R^2$ index for fitting accuracy and Bayesian Information Criterion (BIC) for model selection based on data likelihood and parameter count \cite{neath2012bayesian,vrieze2012model}. Parameter estimation was performed on the Lumosity dataset, and the optimal model results are summarized in Table 3.

Our model outperforms most comparative models in terms of goodness of fit. It is comparable to the second-best model in Flexibility, Language, and Reasoning skills, but excels in Attention, Memory, and overall data performance. While slightly underperforming in Math skill compared to a specific model, our model shows a significant advantage in BIC comparison with six other models due to its smaller parameter size and higher fit level. These results indicate that the auto-discovered symbolic patterns have certain advantages in terms of simplicity and data fitting. This may suggest that in the realm of pattern discovery, AI algorithms may possess certain advantages over humans and have the potential to become important tools in related research.

\section{Discussion}

This paper presents a new method for researchers to identify hidden patterns in data, offering evidence and insights for related studies. A deep regressor is integrated into the model development process to streamline the discovery of symbolic models. The deep regressor, incorporating deep learning architecture (Transformer-like structure) and the IRT model, may include researchers' prior assumptions. Empirical findings show that using deep learning regressors allows for accurate computation of latent states and decreases the search space of symbolic regression. However, the inclusion of priors does constrain the model's generalizability to some extent.

This study used a heuristic model selection approach based on researchers' understanding to establish model selection in symbolic regression. While no universal method exists, exploring universal and rational model selection methods is crucial. In addition, the methodology is limited to continuous feedback scenarios and does not cover discrete feedback. Our future exploration will focus on this, noting that discrete feedback may introduce more noise and make it more difficult.

Finally, this paper examines skill acquisition in psychology research using big data and artificial intelligence. By analyzing core patterns through large-scale NODS, we provide new evidence and methods. However, like other data-driven studies, our research may be influenced by data noise and biases, potentially affecting the generalizability of findings. We recognize the challenges in uncovering learning patterns and acknowledge that our study is based on observations from the Lumosity dataset. Further research is needed to validate generalizability and underlying mechanisms.

\section{Method}
\subsection{Deep learning regression for training log data}
\subsubsection{Problem formulation}
Training log data ($\mathcal{D}$) includes valuable information such as the user set $\mathcal{U}=\{u_1,u_2,…,u_m \}$, the exercise set $\mathcal{E}=\{e_1,e_2,…,e_n\}$, and the skill set $\mathcal{S}=\{s_1,s_2,…,s_k\}$. A user's training history, denoted as $\mathcal{P}_{u_i}=\{p_{u_i}^1,p_{u_i}^2,…,p_{u_i}^l\}$, is a sequence of exercises, where $u_i$ refers to the user, $p_{u_i}^l$ is the $l$-th exercise of $u_i$ and $p_{u_i}^l \in \mathcal{E}$. The sequence $\mathcal{P}_{u_i}$ is obtained by sorting the training history of user $u_i$ by time. The corresponding grade for $\mathcal{P}_{u_i}$ is usually normalized and can be denoted as $G_{u_i}=\{g_{u_i}^1,g_{u_i}^2,…,g_{u_i}^l\}$, where $g_{u_i}^l$ is the grade for the $l$-th exercise of user $u_i$ and $g_(u_i)^l \in [0,1]$. The index matrix indicating the relationship between the exercises and skills is usually called the Q-matrix where $Q \in \mathbb{R}^{n \times k}$. To simplify our notation system, we denote $Q[p_{u_i}^l]$ as the skill index for exercise $p_{u_i}^l$, with the skill index being represented by one-hot encoding on skills, where $Q[p_{u_i}^l ] \in \mathbb{R}^k$.

Exercise scores can generally reflect users' level of skill mastery, and exploring the evolution of these scores can reveal changes in knowledge acquisition. By correlating these scores with practice behavior, the skill acquisition patterns of learners can be determined. Observing the skill acquisition patterns of numerous learners may reveal underlying governing laws that involve psychological activities like practice, forgetting, and skill transfer. An autoregressive self-supervised paradigm is introduced to capture the core patterns of data. Its main objective is to construct a function \textit{f} which predicts the score of any training behavior $p_{u_i}^l$ using its previous sequence of practice sessions, namely, $f({p_{u_i}^1,p_{u_i}^2,…,p_{u_i}^{l-1} })\approx g_{u_i}^l$. If $f$ can predict equivalent scores accurately, it is believed that there exist core patterns or rules about skill acquisition within $f$. To construct $f$, a deep learning model have devised which is inspired by the Transformer architecture. The entire score prediction process is divided into three primary stages: feature encoding, skill mastery inference, and prediction.

\subsubsection{Feature encoding}
The core task of the Feature Encoding module is to extract the features of the historical exercise sequence. Our method involves characterizing the static features of each training behavior, building a model to predict the contribution level of each practice session to corresponding skill learning, weighting the static feature sequence, and obtaining the feature encoding of the practice behavior sequence.

Our study aims to develop the user's practice behavior from three dimensions, namely, user attributes, exercise characteristics, and scheduling information. User attributes typically consist of personal details such as age, occupation, and education, which can offer valuable information about motivation, initial skill level, and learning speed that could impact skill learning. We will denote $u_i$'s user attributes as $\textbf{\textit{U}}[u_i]$. Exercise characteristics aim at conveying relevant aspects of the exercise, notably the relationship between the exercise and the skill, and the kind of stimulus incorporated in the exercise. We will use the notation $\textbf{\textit{E}}[p_{u_i}^{l-1}]$ to denote the exercise characteristics of exercise $p_{u_i}^{l-1}$. Scheduling information comprises details about the user during the exercise process, such as the number of exercises intended to learn a skill and the time between successive exercises. The scheduling information pertains to the data about the user's actions and behavior during the ongoing exercise, which is subsequently used for prediction. We will refer to the scheduling information of user $u_i$ during exercise $p_{u_i}^l$ by using the notation $\textbf{\textit{S}}[p_{u_i}^l ]$. Typically, features can take on two main forms, namely, continuous and discrete. For discrete features, we employ one-hot encoding for encoding purposes, while for continuous variables, we directly use their continuous values as feature values. In the end, the static features of a user's practice behavior can be documented as $\textbf{\textit{SF}}[p_{u_i}^{l-1}]=\textbf{\textit{U}}[u_i] \oplus \textbf{\textit{E}}[p_{u_i}^{l-1}] \oplus \textbf{\textit{S}}[p_{u_i}^l]$, where $\oplus$ means the concatenation operator.

Since different training behaviors may have distinct impacts on skill learning, we employ neural networks to determine the significance of specific behaviors for skill acquisition. First, to normalize the static features and expedite model training while enhancing stability, we perform batch normalization, given that the distinct dimensions of the static features have dissimilar value ranges. Second, we employ a classical multilayer perceptron (MLP) to model normalized features, where the perception machine varies depending on the skill but shares the parameters for the same perception machine. Consequently, for a user's practice behavior $p_{u_i}^{l-1}$ related to skill $s_k$, we can quantify its importance as
\begin{equation}
	\alpha_{p_{u_i}^{l-1}}^{s_k}=\mathrm{MLP}_{att}\left( \mathrm{BN}\left( \textbf{\textit{SF}}[p_{u_i}^{l-1}]\right) ,\bm{\Theta}_ {att}^{s_k}\right) 
\end{equation}
where MLP and BN are abbreviations for multilayer perceptron and batch normalization, respectively. We will provide detailed information on the methodology in the "Method" section. $\bm{\Theta}_ {att}^{s_k}$ refers to the set of MLP parameters.

The user attributes, which make up one of the three static features, refer to time-invariant variables and do not necessitate temporal feature fusion. Since scheduling information inherently encompasses temporal information during encoding, the temporal encoding of user exercise features becomes the primary focus of sequence encoding. The encoded exercise feature of $u_i$’s exercise sequence ${p_{u_i}^1,p_{u_i}^2,…,p_{u_i}^{l-1}}$ on skill $s_k$ can be represented by the following equation:
\begin{equation}
	\textit{\textbf{EE}}^{s_k}[\{p_{u_i}^1,p_{u_i}^2,…,p_{u_i}^{l-1} \}]=\sum_{j=1}^{l-1} \alpha_{p_{u_i}^j}^{s_k}\cdot \textit{\textbf{E}}[p_{u_i}^j]
\end{equation}
The encoded behavioral sequence feature for skill $s_k$ can be expressed as: $ \bm{\omega}^{s_k} [{p_{u_i}^1,p_{u_i}^2,…,p_{u_i}^{l-1}}]=\textit{\textbf{U}}[u_i]\oplus \textit{\textbf{EE}}^{s_k} [{p_{u_i}^1,p_{u_i}^2,…,p_{u_i}^{l-1}}]\oplus \textit{\textbf{S}}[p_{u_i}^l]$.

\subsubsection{Mastery inference}
Mastery inference focuses on constructing a mapping function (mastery inferer) that maps encoded features of a behavioral sequence to the level of skill mastery, which is the central component of the model. As the form of the function is ambiguous, we propose utilizing a neural network model to fit the mapping function and leverage its formidable non-linear fitting ability to precisely estimate the skill mastery variable. Additionally, score prediction necessitates employing another mapping function that translates the skill mastery state into the final score. To this end, we are going to incorporate the classic cognitive diagnostic model (item response theory \cite{IRT3,IRT4}) as the mapping function.

The mastery inferer for each skill is independent. Therefore, we set up a group of neural network models with the same structure but different parameters to estimate the level of mastery for each skill. When user $u_i$ practices $p_{u_i}^l$, the level of mastery for skill $s_k$ can be denoted as $\phi_{p_{u_i}^l}^{s_k}$, and it can be expressed as:
\begin{equation}
	\phi_{p_{u_i}^l}^{s_k}=\psi_{u_i}^{s_k}+\mathrm{MLP}_{msty} (\mathrm{BN}(\bm{\omega}^{s_k} [{p_{u_i}^1,p_{u_i}^2,…,p_{u_i}^{l-1}}]),\bm{\Theta}_ {msty}^{s_k} )
\end{equation}
where MLP and BN are short for multilayer perceptron and batch normalization, respectively. We represent the level of mastery of each skill of user $u_i$ by $\bm{\Phi}_{p_{u_i}^l}$, where $\bm{\Phi}_{p_{u_i}^l}=[\phi_{p_{u_i}^l}^{s_1},\phi_{p_{u_i}^l}^{s_2},…,\phi_{p_{u_i}^l}^{s_k}]$ and $s_1,s_2,…,s_k\in \mathcal{S}$. $\psi_{u_i}^{s_k}$ denotes the initial proficiency level of skill $s_k$ possessed by user $u_i$ before training.

\subsubsection{Score prediction}
Generally, a user's score on an exercise is believed to be correlated with their level of skill mastery, which has been extensively studied by scholars who have developed numerous cognitive diagnostic models. In this paper, we propose using the item response theory (IRT) model \cite{IRT3,IRT4} as the scoring prediction function. Specifically, we introduce a three-parameter IRT model that models user mastery, exercise difficulty, discrimination, and guessing, enabling us to better estimate exercise scores using user mastery information. Its specific form can be expressed as:
\begin{equation}
	\hat{g}_{u_i}^l=\gamma_{p_{u_i}^l}+\frac{1-\gamma_{p_{u_i}^l}}{1+exp\left( -d\cdot\eta_{p_{u_i}^l}\cdot\left( \bm{\Phi}_{p_{u_i}^l}-\beta_{p_{u_i}^l}\cdot Q[p_{u_i}^l ]\right) \right) }
\end{equation}
Here, $\eta_{p_{u_i}^l}$, $\beta_{p_{u_i}^l}$ and $\gamma_{p_{u_i}^l}$ respectively denote the discrimination, difficulty, and guessing of the exercise $p_{u_i}^l$. $Q[p_{u_i}^l]$ represents the index of the exercise in the skill set, while $\hat{g}_{u_i}^l$ represents the predicted score for practice $p_{u_i}^l$.

The neural network model used in this article is implemented with the open-source deep learning library Pytorch\cite{Pytorch}.

\subsubsection{Model learning}
In the preceding section, we denoted $\hat{g}_{u_i}^l$ as the forecasted score for exercise $p_{u_i}^l$, obtained through the feature encoding, transformation and forecasting process from the learning sequence $\{p_{u_i}^1,p_{u_i}^2,…,p_{u_i}^{l-1} \}$. Firstly, since scoring prediction is a typical regression task, we introduce mean absolute error (MAE) loss function as our optimization objective. MAE is known to have better resistance against outliers as compared to the more commonly used mean squared error (MSE) loss. Our main objective is to extract knowledge that is universally applicable and generalizable from the data. To accomplish this objective, we endeavor to decrease the effect of score values that are unusual or stand out on the model. Secondly, we intend to adopt a self-regressive learning paradigm. Therefore, the established loss function needs to be computed and accumulated on all the practice session data of every learner in the dataset.

Considering the significance of user training behavior, shown in Eq. (1), we seek to incorporate a constraint that ensures smoothness. The smoothness constraint serves to restrict the importance values to a predefined range. If the model is capable of constraining all the important values to a constant value without compromising precision, it can diminish the overall complexity of the model and alleviate the difficulty of interpreting it.

The learning objective function of the entire model can be represented as follows.
\begin{equation}
	\mathcal{L}=\underbrace{\frac{1}{\left| \mathcal{D} \right| } \sum_{u_i \in \mathcal{U}}\sum_{p_{u_i}^l \in \mathcal{P}_{u_i}}\left| \hat{g}_{u_i}^l-g_{u_i}^l \right| }_{\text{MAE loss}} + \lambda \cdot \underbrace{\frac{1}{\left| \mathcal{D} \right| }\sum_{u_i \in \mathcal{U}}\sum_{p_{u_i}^l \in \mathcal{P}_{u_i}} \frac{1}{\left| \mathcal{S}\right| } \sum_{s_k \in \mathcal{S}} \left(\alpha_{p_{u_i}^l}^{s_k}-1  \right)^2  }_{\text{regularization term}}
\end{equation}
The objective function above contains two main components. The first component is the fidelity term, which describes the discrepancy between the predicted data and the actual data, represented by the MAE loss. The second component is the regularization term, namely, the smoothness constraint on the importance of training behavior, which is confined to a constant value near 1. $\lambda$ is a balancing factor used to balance the importance of the fidelity term and the constraint term. $\left| \bullet\right| $ means the cardinality of a set. To be more specific, $\left| \mathcal{D} \right|$ represents the number of training records in the dataset, where $\left| \mathcal{D} \right| = \left| \mathcal{U} \right| \times \left| \mathcal{P}_{u_i} \right|$. Since the entire model is differentiable, it is capable of conducting efficient learning of parameters through gradient-based iterative optimization algorithms, for instance, stochastic gradient descent (SGD), and adaptive moment estimation (Adam) \cite{adam}.

\subsection{Symbolic law extraction from deep learning regressor}
\subsubsection{Symbolic distillation}
Formally, let \textit{T} be the neural network model to be analyzed, and let \textit{S} be the symbolic regression model. The core idea of symbolic distillation is to ensure that for any input \textit{X}, $T(X)\approx S(X)$. Based on this, we can define the optimization objective for the symbolic regression model as follows
\begin{equation}
	\pi=\mathop{\arg\min}\limits_{\pi}\mathbb{E}_{x\sim X}\left| T(x,\bm{\Theta})-S(x,\pi)\right| 
\end{equation}
where $\pi$ represents the algebraic equation learned by the symbolic regression model, x represents the sampled input feature, $\bm{\Theta}$ represents all the parameters in the neural network model, which could be considered constants after training.

Specifically, in this article, we construct $X$ using the input information samples from the training data. In order to reduce the computational cost of the symbolic regression model, we sample the input samples from the entire dataset to obtain $X$, where $X\in \mathbb{R}^{s\times k}$. Here, $s$ represents the sampling size of the samples, which is a hyperparameter set manually, while $K$ represents the feature dimension of the neural network input information, i.e. the dimension of $\textbf{\textit{SF}}[p_{u_i}^{l-1}]$ or $\bm{\omega}^{s_k} [{p_{u_i}^1,p_{u_i}^2,…,p_{u_i}^{l-1}}]$, as shown in Eqs. (1) and (3).

In our study, features have a high dimensionality. Symbolic regression is a typical combinatorial optimization problem. When the number of variables increases, the solution space grows exponentially, diminishing the efficiency and stability of the symbolic regression model. As an exploratory study, our objective is to identify core patterns and essential factors. Consequentially, we use a trained neural network model and an interpretable analysis method, based on gradient feature importance estimation, to screen features, select the most important features, and use them to execute the symbolic regression to identify key patterns. Specifically, the trained neural network model and the gradient feature importance estimation method are incorporated to calculate the importance of the $K$ features, among which the k most crucial ones are chosen to form the input feature $C(X)$. Here, $C(X)\in \mathbb{R}^{s\times k}$, and $k\ll K$. We will describe the detailed gradient-based feature importance estimation method calculation in the subsequent Method section. In addition, the neural network modules in the deep learning regressor (attention-based encoders and mastery inferers for each skill, shown in Fig. 1b) consist of two operators, namely BN and MLP. To avoid feature scaling impact on importance estimation, we restrict our feature importance estimation to the output features of the BN operators.

Thus, the objective of symbolic regression is formulated here as follows: 
\begin{equation}
	\pi=\mathop{\arg\min}\limits_{\pi}\mathbb{E}_{x\sim X}\left| T(x,\bm{\Theta})-S(C(x),\pi)\right| 
\end{equation}
In order to avoid the over-influence of errors on the model, mean absolute error is introduced as the objective function for the symbolic regression model. Genetic algorithms are the primary approach employed to solve the symbolic regression model, symbolic regression libraries such as Eureqa, PySR, and geppy offer additional means of solving the model. The symbol regression algorithm used in this article is implemented with the open-source symbol regression library PySR.

\subsubsection{Governing laws extraction}
In this section, we are going to depict how we convert the prediction process of trained deep regressor into a symbolic law using the symbolic distillation method. We express the neural network models in each module of the deep regressor as a symbolic law and ultimately combine them to obtain the symbolic representation of the entire deep regressor. As the attention-based encoder and mastery inferer for each skill are independent, it is feasible to perform symbolic distillation on these components in parallel. Furthermore, since the attention-based encoder and mastery inferer are coupled and exhibit sequential order, the expansion process of the analysis follows a similar order. First, for the attention-based encoder, a symbolic representation of $\alpha_{p_{u_i}^l}^{s_k}$ is obtained by symbolically analyzing equation (1), merged with feature encoding (equation (2)), producing a symbolic representation of the encoded features, i.e., $\bm{\omega}^{s_k} [{p_{u_i}^1,p_{u_i}^2,…,p_{u_i}^{l-1}}]$. If the model includes a smoothness constraint for attention value, $\alpha_{p_{u_i}^l}^{s_k}$ can be approximated to 1 when the smoothness component's value is less than an extremely small threshold (e.g., 1e-4) to reduce the complexity of the analysis equation. Next, the mastery inferer undergoes a symbolic distillation, followed by the integration of symbolic representation with the encoded features, reorganized to render the final symbolic governing laws.

\subsection{Data generation method and model parameter setting in the simulated data experiment}

To closely simulate real exercise patterns, three commonly accepted laws of skill acquisition were considered: Power law ($\beta \cdot {N}^\alpha$ ), Exponential law ($\beta \cdot \exp(\alpha \cdot N)$),and Linear law ($\beta \cdot \alpha \cdot N$). The concept of mutualism, indicating the positive impact of exercising on multiple skills, was also integrated, where mastering skill A could enhance proficiency in skill B. Additionally, a forgetting factor and proficiency degradation model based on exercise interval were introduced. Three sets of simulated skill mastering laws were established, detailed in Table 1.

To generate comprehensive simulated training log data, learners and items were first simulated with randomly initialized parameters. Subsequently, each learner was assigned a random exercise sequence. By considering the exercise process and pre- determined learning patterns, skill mastery for each exercise was calculated for the learners. The IRT model, along with item parameters, was utilized to compute scores for each exercise. Gaussian noise ($\varepsilon\sim N\left( 0,\delta \right)$) was introduced during the calculation of practice scores to assess the algorithm’s robustness to noise and create data that mirrors real-world observations \cite{errorform,dataset1}.

The simulated training log data generation can be divided into three main stages: learners and exercises generation, learning path generation, and cognitive state and learning feedback generation, shown in Extended Data Figure 2. In the learner simulation, a total of 50 independent virtual learners are simulated. In the practice simulation, 20 simulated test questions are generated for each of skill 1 and skill 2, each test question containing three coefficients: difficulty, guessing, and discrimination, all of which are random variables uniformly distributed between 0 and 1. Learning path generation involves generating the practice sequence for each simulated learner. To simplify the simulation environment setup process, the corresponding learning sequence is generated by randomly selecting questions, with each simulated learner's learning sequence length being 20. Combining the learner's learning path with the predetermined skill acquisition function (in Table 1), the learner's skill mastery can be calculated at the beginning of each practice. Finally, combining the parameters of the test questions, the learner's test score is calculated using the three-parameter IRT model (Eq. 4) {and Gaussian random noise with a mean of 0 and a variance of $\delta$ is added.}

Under the three settings, the parameters of the Linear law are $\alpha$=1, $\beta$=0.3, $\gamma$=0.1, the parameters of the Exponential law are $\alpha$=0.1, $\beta$=1, $\gamma$=0.1, and the parameters of the Power law are $\alpha$=0.8, $\beta$=1, $\gamma$=0.2. {In the three settings considered, the mutualism factor $\mu$ is set to 0.5. In the simulation experiment, four different levels of noise are selected as test standards, namely $\delta\in\left\lbrace0,0.001,0.01,0.1 \right\rbrace $.} In the deep learning regressor, the MLP models in feature encoding and mastery inference are both three-layered (two hidden layers and one output layer), with 1000 neurons in each hidden layer and Leaky ReLU as the activation function. The model optimizer is set to Adam, and the learning rate is 1e-3. In the symbolic regression model, the number of iterations is set to 2000, the population size is 100, and the maximum function length is 15. The preset operators include "+", "-", "$\times$", "pow", "exp", "log", and "sin".

\subsection{Data preprocessing method and model parameter setting in the real-world data experiment}
Lumosity (www.lumosity.com) is an online platform that offers brain training games and exercises designed to improve cognitive functions such as memory, attention, flexibility, speed of processing, and problem-solving. The platform was created by Lumos Labs and is based on the concept of neuroplasticity \cite{lumos}, which is the brain's ability to change and adapt throughout a person's lifetime. Lumosity has been used by millions of people worldwide to enhance their cognitive abilities and improve their overall mental fitness. In the platform, users improve their proficiency in corresponding skills by playing various games, which can be considered as cognitive training or practice as mentioned earlier. By analyzing the extensive practice sessions of numerous users, we aim to uncover the behavioral patterns of skill acquisition hidden behind the data.

The dataset used in this study was shared by \cite{dataset1}. We performed data screening and preprocessing based on the raw data of log data containing 50 million exercises provided by them. Finally, we selected 6143 users as research sample and extracted their first 2000 practice logs as the dataset. Secondly, based on the data labeling foundation of \cite{dataset1}, we identified 37 features as behavioral features, including all U, E, and S type features, and the detailed feature encoding, description, type, and abbreviation can be found in Extended Data Figure 3. In the dataset, there are a total of 85 games, and each game is labeled with its corresponding skills and subskills. Skills includes Attention, Flexibility, Language, Math, Memory, and Reasoning, which are the dimensions we considered when estimating users' skill mastery. Subskill represents more fine-grained skill types, which are included in skill, and the relationship between them can be found in Extended Data Figure 3. In the experiment, Subskill only serves as an input feature of the model and is not the prediction target.

In this paper, we utilized raw data provided by \cite{dataset1}. To ensure the quality of our data, we filtered out games with fewer contact times and related skills. We then selected the first 2000 practice records of users with more than 2000 practice times as our training data. To avoid any issues with abnormal scores, we normalized the score data of each game for the selected training data. Specifically, we took the top 10\% of scores as the full score and performed maximum-minimum normalization on the data of each game. All scores were then uniformly scaled to a range of [0.01,0.8] to prevent the Inf outlier problem caused by the cognitive diagnostic model.

The deep learning regressor employs three-layered MLP models for feature encoding and mastery inference, each consisting of two hidden layers and one output layer. The activation function used is Leaky ReLU, and each hidden layer comprises 1000 neurons. The model optimizer is set to Adam, with a learning rate of 1e-4. The symbolic regression model is configured with a population size of 100, a maximum function length of 15, and a fixed set of operators, including "+", "-", "*", "pow", "exp", "log", and "sin". The model is trained for 2000 iterations to achieve optimal performance. To increase the interpretability of governing laws, operators other than "+", "-", and "*" are not allowed to be nested. In the symbolic regression model, a total of 2000 training log samples are selected randomly, and the most important 8 features are screened through the feature importance analysis method. To increase the interpretability of governing laws, operators other than "+", "-", and "$\times$" are not allowed to be nested.

\bmhead{Data availability}
All data generated or analyzed during this study are available. It can be found in https://github.com/ccnu-mathits/ADM. And it is also available via Zenodo at https://doi.org/10.5281/zenodo.10938670 (ref. \cite{code}). Source data are provided with this paper.

\bmhead{Code availability}
The source code of this study is freely available on Github at https://github.com/ccnu-mathits/ADM. And it is also available via Zenodo at https://doi.org/10.5281/zenodo.10938670 (ref. \cite{code}).

\clearpage

\appendix
\setcounter{figure}{0}
\renewcommand{\figurename}{Extended Data Fig.}

\begin{figure}[p!]%
	\centering
	\includegraphics[width=0.9\textwidth]{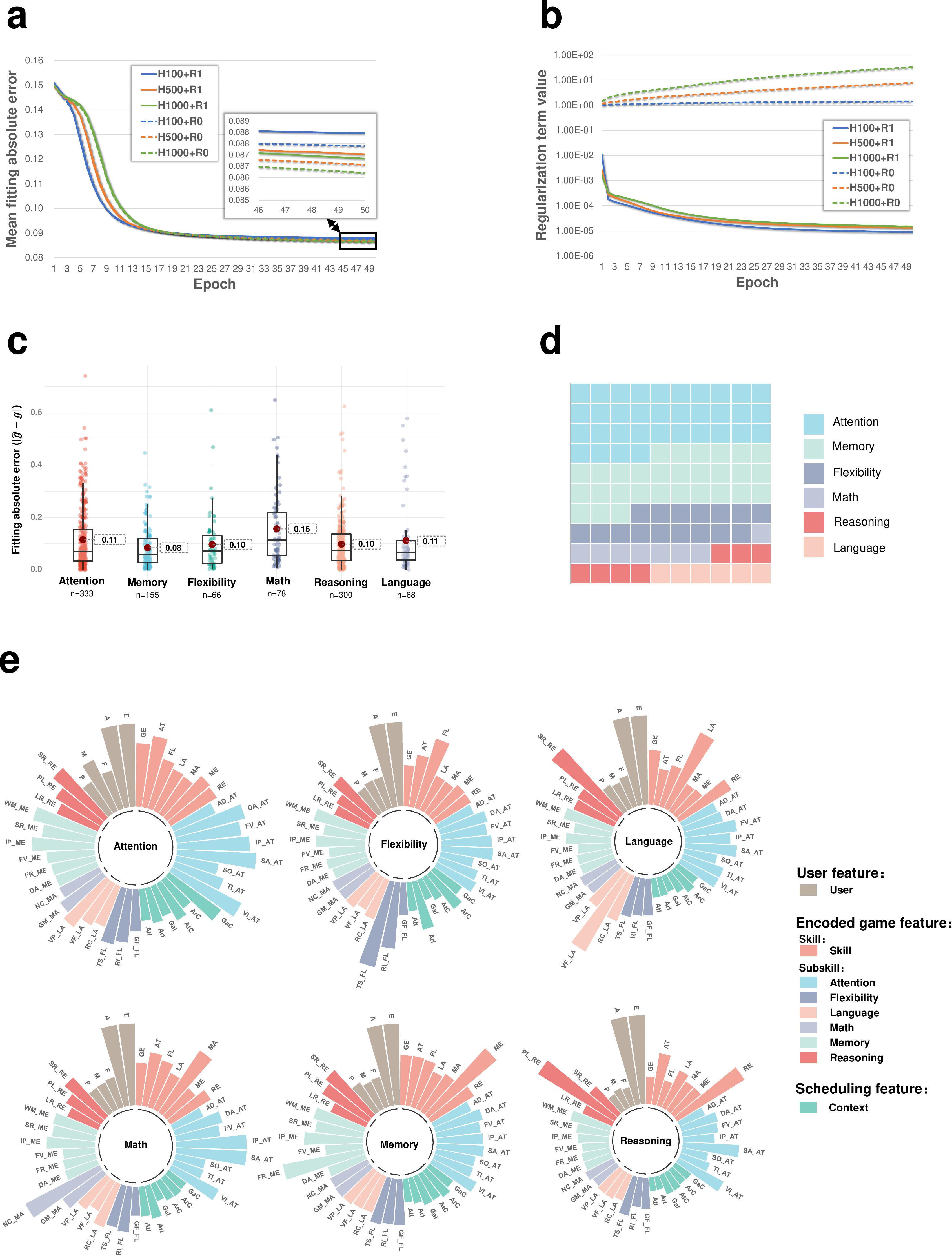}
	\caption{Partial results on the Lumosity dataset. More specifically, (a) the change curve of mean fitting absolute error during the iteration process of the deep learning regressor; (b) the change curve of the value of the regularization term during the iteration processes; (c) the prediction error distribution of 1000 randomly selected records from the trained H1000+R1 model. The box plot displays the interquartile range (IQR) with the median line, while the whiskers extend to the minimum and maximum values or a multiple of the IQR from the quartiles. Outliers are depicted as individual points beyond the whiskers; (d) the proportion of the number of practice times for each skill to the total number of practice times; (e) the distribution of feature importance for each skill in H1000+R1.}\label{efig1}
\end{figure}

\begin{figure}[p!]%
	\centering
	\includegraphics[width=0.9\textwidth]{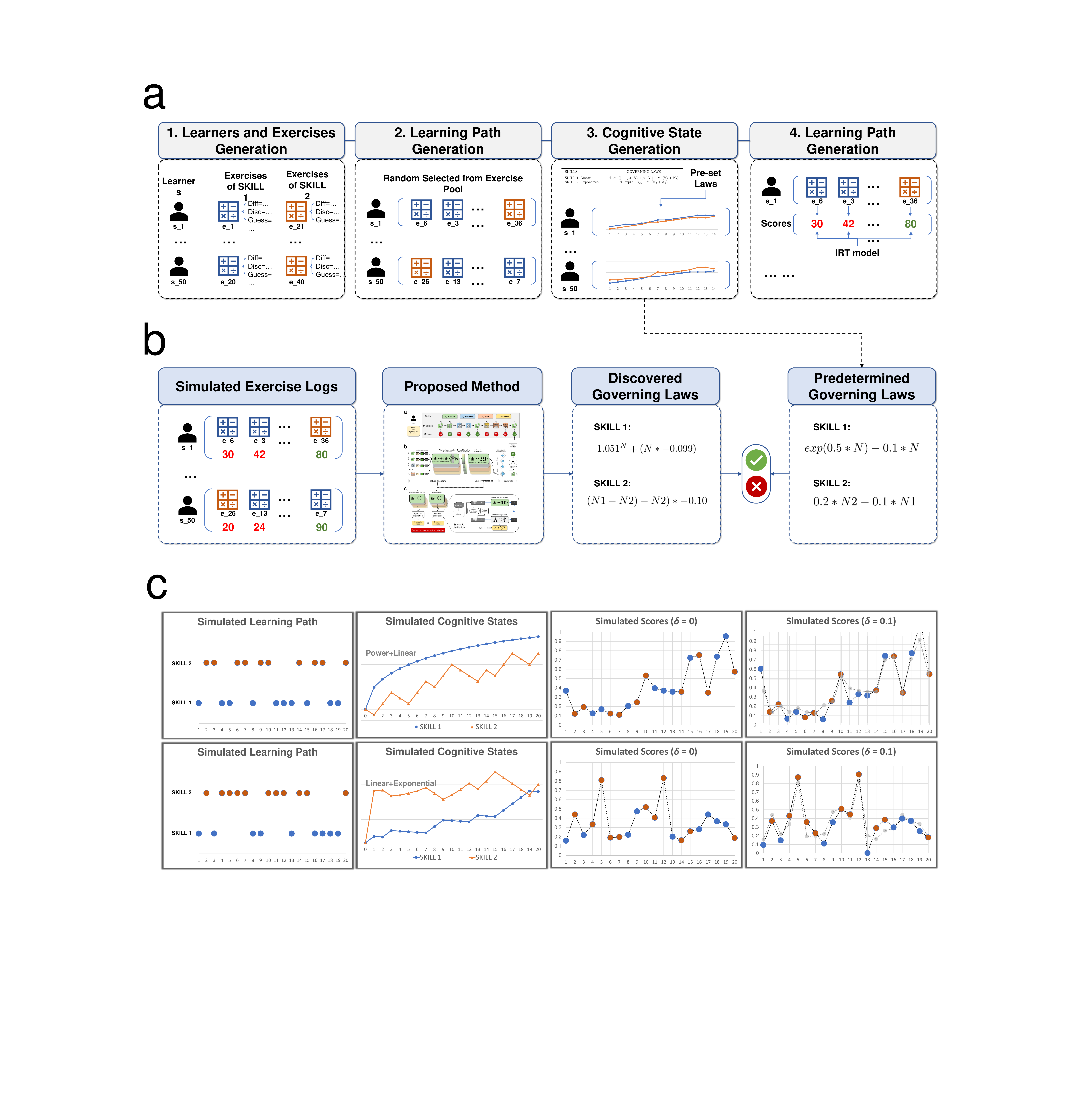}
	\caption{Schematic representation of the simulation experiment. (a) Flowchart depicting the process of generating simulated data and the data format. (b) Schematic diagram illustrating the evaluation procedure of the simulation experiment. (c) Sample representation of the simulated data.}\label{efig2}
\end{figure}

\begin{figure}[p!]%
	\centering
	\includegraphics[width=0.9\textwidth]{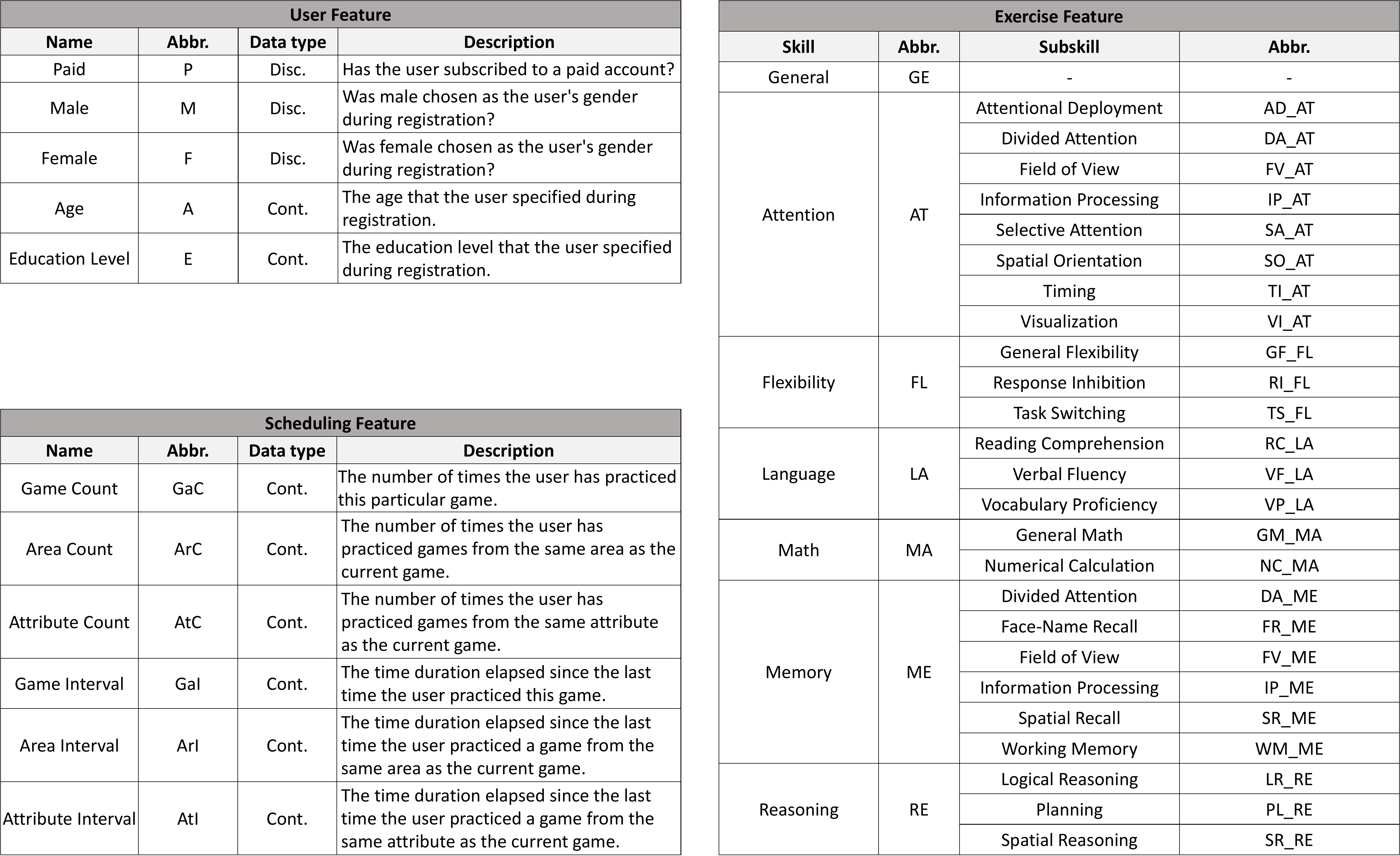}
	\caption{The feature set utilized in the real-world dataset (Lumosity) experiment, which comprises three components: user (U), exercise (E), and scheduling (S). The features consist of two categories: discrete and continuous. Discrete features are encoded through one-hot encoding, while continuous features are encoded using real values. Game features are two-dimensional, consisting of skill and subskill, both of which are discrete features and characterize the relationship between the game and cognitive skills. Subskill is a subdivision feature of skill.}\label{efig3}
\end{figure}

\begin{figure}[p!]%
	\centering
	\includegraphics[width=0.9\textwidth]{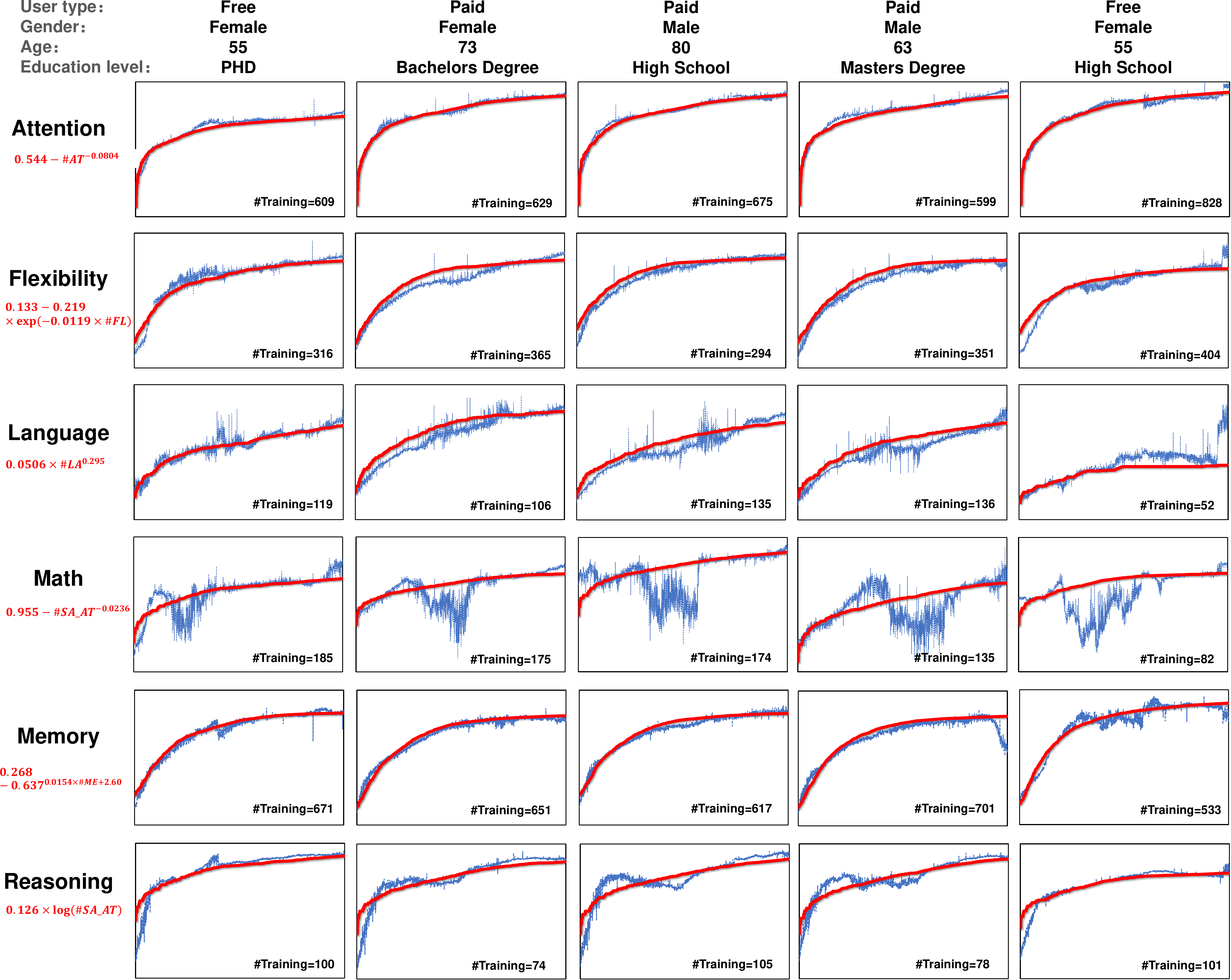}
	\caption{Analysis of the difference between the mastery level of skills calculated by the deep learning regressor and the discovered governing laws. Here, we present the changes in skill proficiency at each time point during 2000 practice sessions for five learners. The blue line represents the mastery curve computed by the deep regressor, while the red line represents the curve computed by the discovered governing law. The governing law is the same for all learners, but due to differences in their choices and order of practice, there are variations in the independent variable of the governing law for each learner.}\label{efig4}
\end{figure}

\begin{figure}[p!]%
	\centering
	\includegraphics[width=0.9\textwidth]{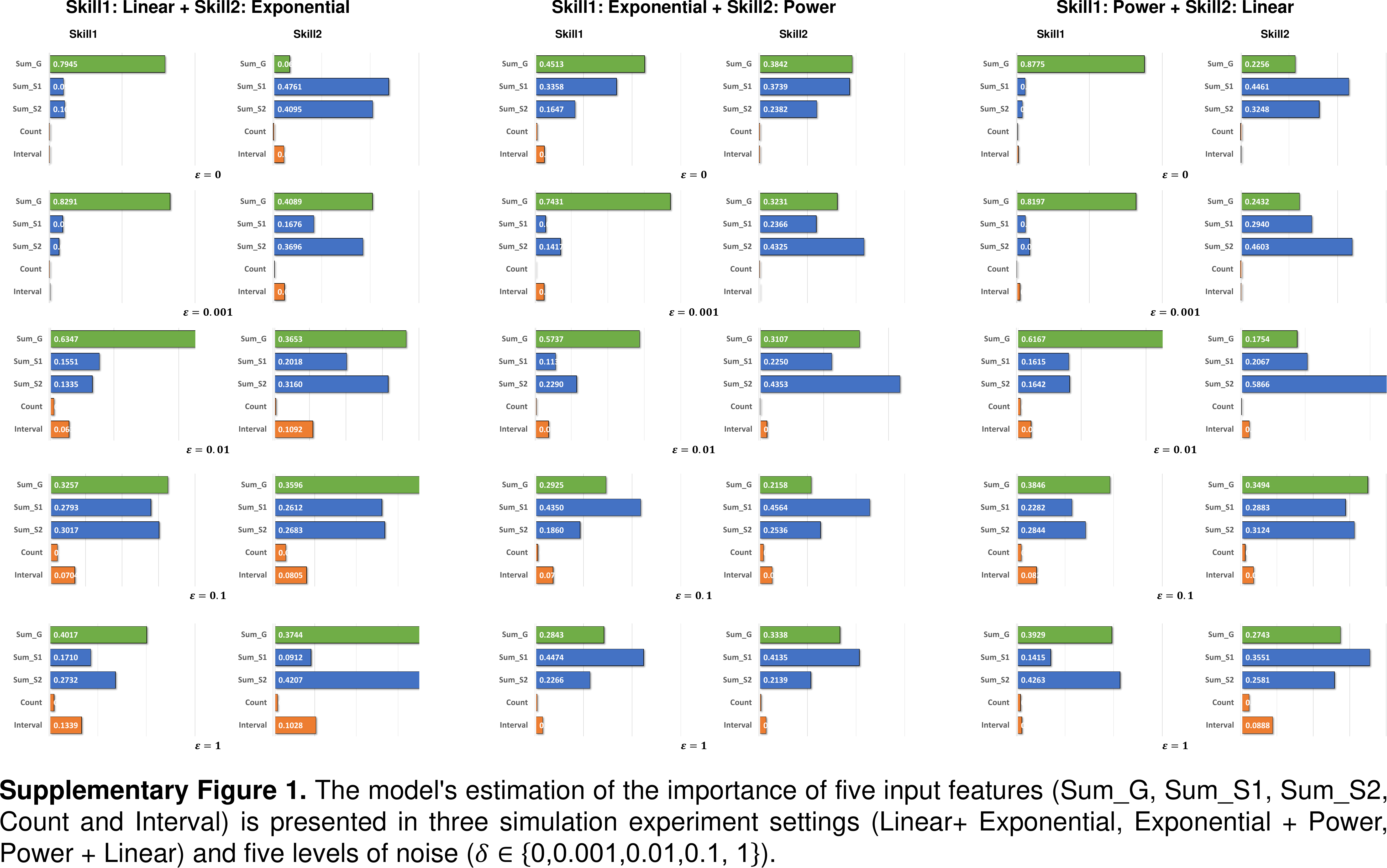}
\end{figure}
\begin{figure}[p!]%
	\centering
	\includegraphics[width=0.9\textwidth]{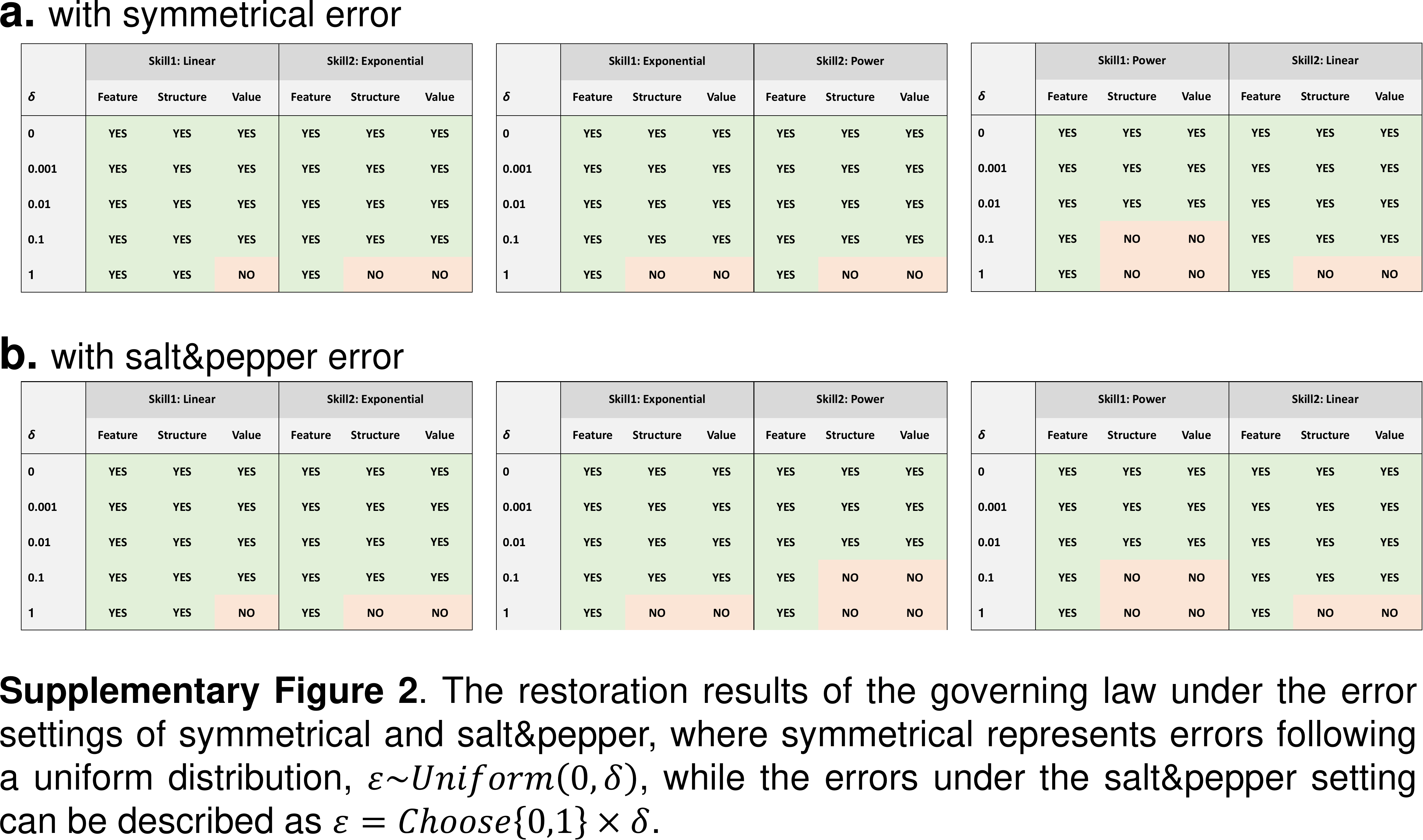}
\end{figure}

\clearpage

\end{document}